\documentclass[twoside,11pt]{article}

\usepackage{hyperref}
\usepackage{svg}
\usepackage{jair, theapa, rawfonts}
\usepackage{ragged2e}
\usepackage{enumitem}
\usepackage{linguex}
\usepackage{multirow}
\usepackage{booktabs}
\usepackage{array}
\usepackage{tablefootnote}
\usepackage{amsmath}
\usepackage{xcolor}
\usepackage{graphicx}
\usepackage{acronym}

\firstpageno{1}

\begin{document}

\title{Neural Conversation Models and How to Rein Them in:\\
A Survey of Failures and Fixes}

\author{\name Fabian Galetzka \email fabian.galetzka@volkswagen.de \\
       \addr Computational Linguistics\\ University of Potsdam\\
       \AND
       \name Anne Beyer \email anne.beyer@uni-potsdam.de \\
       \addr Computational Linguistics\\ University of Potsdam\\
       \AND
       \name David Schlangen \email david.schlangen@uni-potsdam.de \\
       \addr Computational Linguistics\\ University of Potsdam\\}


\maketitle

\begin{abstract}
Recent conditional language models are able to continue any kind of text source in an often seemingly fluent way. This fact encouraged research in the area of ``open-domain'' conversational systems that are based on powerful language models and aim to imitate an interlocutor by generating appropriate contributions to a written dialogue. From a linguistic perspective, however, the complexity of \emph{contributing to a conversation} is high. In this survey, we interpret Grice's maxims of cooperative conversation from the perspective of this specific research area and systematize the literature under the aspect of what makes a contribution \emph{appropriate}: A neural conversation model has to be fluent, informative, consistent, coherent, and follow social norms. In order to ensure these qualities, recent approaches try to ``tame'' the underlying language models at various \emph{intervention points}, such as data, training regime or decoding. Sorted by these categories and intervention points, we discuss promising attempts and suggest novel ways for future research.
\end{abstract}

\vspace{.5cm}

\noindent This review was started in 2021 and was largely finalised in 2022. The present version represents a lightly edited variant of a chapter from the first author's PhD thesis~\cite{Galetzka2022}. Note that the article hence does not cover any material newer than 2022. We believe, however, that at least the systematic presentation of the challenges still has value, even if in the current state of the art, not all of the intervention points are still equally accessible.

\section{Introduction}
\label{sec:intro}


With the advent of powerful conditional language models, the idea of building ``open-domain'' conversational systems that can simulate casual conversation about anything and everything has seen a resurgence as a respectable research goal (rather than being left to the pursuit of hobbyists).

Two recent developments have contributed to this. First, with the increasing popularity and use of social media, ever larger amounts of conversational data (e.g., posts and questions and answer threads) have become easily available. \shortciteA{ritter2011data} was an early example of how these data can be made useful for the task of modelling conversations; in this case using techniques from statistical machine translation to ``translate'' utterances into responses to them. In parallel to this, statistical language models (and therefore conditional language generators) have become more powerful due to new architectures like Self-Attention \shortcite{NIPS2017_3f5ee243}, an increase in raw computational power, and, with this, an increase in the size of the trainable models.

Early intuitions suggested that these ingredients might be enough to build open domain conversational systems. Even though first experiments with such models  yielded surprisingly fluent and seemingly fitting responses given short contexts, these hopes were quickly dashed with the subsequent observation of many problems, such as an overuse of uninformative answers (\textit{``I don't know''}), inconsistent personality projections \shortcite{vinyals-etal-2015-a-neural-conv}, or generally missing context awareness \shortcite{Serban2016BuildingED}. 
After the early enthusiasm, to a certain degree a realisation of the complexity of the task set in, answering, as it does, to a variety of constraints from both outside the discourse (accuracy of factual claims) and inside of it (needing to fit to what has previously been said).

In this paper, we attempt to systematise the literature about the attested problems of \textit{neural conversation models} (conditional language models realised with neural networks) used as chat-partner simulators, and approaches to addressing these. We show that these approaches can be framed as attempts to ``tame'' the underlying model, in the sense that they impose further constraints on it. For this, we identify different targets or \textit{intervention points}, such as data, training regime, decoding, etc. We show that the constraints can be related to observations from the field of pragmatics, and more specifically to Grice's \textit{maxims of cooperative conversation} \shortcite{LogicandConversation}.\footnote{%
\shortciteA{higashinaka-etal-2015-towards} similarly take inspiration from Grice's cooperativity principle in their taxonomy of chat-bot error types, but do not relate these specifically to neural models and attempts at improving those.} 

Finally, we show that this relation can be made productive, by suggesting, for future work, novel ways to address problems.

This paper aims to provide both a historical overview of developments in this field as well as serve as a catalogue of the various ways of improving models that have been explored, and through this, to be of use to researchers looking for ways to improve their models.

\section{Contributing to a Conversation}
\label{ch:3_grice_contribution}
A first step towards understanding what properties a neural conversation model needs to have to carry on any conversation appropriately, we start with analyzing what a contribution to a conversation is first.

\subsection{The Task, Formalized}

The task of participating in a conversation can be seen as identifying the contextually most appropriate contribution $c_{t}^*$ at turn $t$ out of all possible contributions $c$:
\begin{equation}
	c_{t}^* = \mbox{most\_appropriate}_c \; R(c | C_t).
\end{equation}
with $R$ being a conversation model that generates contributions $c$ given a context $C_t$. In order to determine a contextually appropriate contribution, $C_t$ needs to include at least the observable public context, which is the dialogue history $D_t=(o_1, c_1, ...o_t)$. It consists of observations $o_i$ (contributions by the dialogue partner) and own (previous) contributions $c_i$ until the current speaker turn $t$. 

However, the observable context is not sufficient. Additionally, a non-observable internal state is needed, otherwise any latent variables that control the internal consistency within $c_i$ must be fully expressed in the observable information. This dimension of consistency can be achieved by providing additional private information $K$ to the model. $K$ can contain factual information (about the world in which the conversation is taking place) or personal information, for example stylistic parameters (make utterances conform in style to the personality ''40-year-old male accountant'') or proffered information (make self-claims consistent with ''40-year-old male accountant''). The full formulation of the context $C_t$ then is:
\begin{equation}
	C_t = (D_t, K) \mbox{ with }  K = \{p_1, \dots, p_n, f_1, \dots, f_m\},
\end{equation}
where the $p_i$ are the persona facts and the $f_i$ the general facts.

We further need to clarify the exact meaning of ``appropriateness'' to the public and private context $C_t$ at turn $t$. For this, we introduce Grice's model of cooperative conversation in the following Subsection~\ref{ch:3_grice_contribution_coorperative_principle} and adapt it to the research area of neural generation models in Subsection~\ref{ch:3_grice_contribution_appropriateness}.

\subsection{The Cooperative Principle}
\label{ch:3_grice_contribution_coorperative_principle}

The philosopher H.P.\ Grice proposed in 1975 in his seminal paper ``Logic and Conversation'' the adherence to the \emph{Coorperative Principle} as a fundamental feature of contribution to discourse: 

\begin{quote}
	Make your conversational contribution such as is required, at the stage at which it occurs, by the accepted purpose or direction of the talk exchange in which you are engaged. \hfill \cite[p.\ 45]{LogicandConversation}
\end{quote}

This theoretical proposal attempts to explain certain aspects of discourse meaning, where meaning can also come from apparently violating these maxims. The proposal has been enormously influential in linguistic pragmatics and has also been successfully used in Natural Language Processing (e.g., \shortciteA{heemanhirst:collab}), which motivates us to take it as our starting point as well. He lists nine conversational maxims, sorted into four categories, that need to be followed to make a contribution ``such as is required''.

\begin{figure}
	\centering
	\includegraphics[scale=0.13]{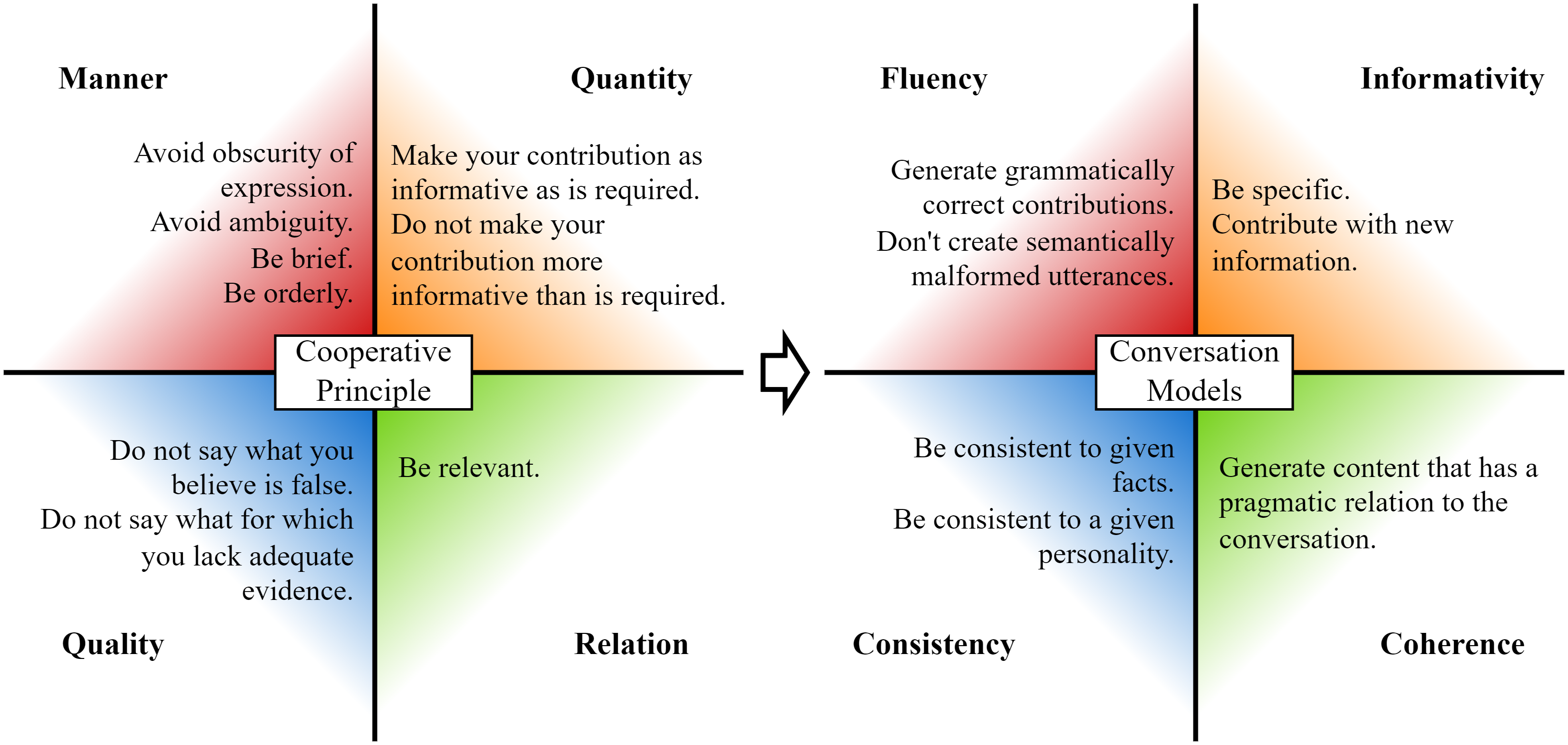}
	\caption{Operationalising the Conversational Maxims \shortcite{LogicandConversation} as Demands on Conversation Models.}
	\label{fig:grice_adapted_to_convLM}
\end{figure}

The category \textbf{Manner} includes the supermaxim `\textit{Be perspicuous}' and relates not to what is said, but how it is said. Grice listed the maxims:
\setlist{noitemsep}
\begin{enumerate}
	\item Avoid obscurity of expression.
	\item Avoid ambiguity.
	\item Be brief.
	\item Be orderly.
\end{enumerate}
The category \textbf{Quantity} relates to the right amount of information that is required in a conversation. Grice formulated the following maxims:
\setlist{noitemsep}
\begin{enumerate}
	\setcounter{enumi}{4}
	\item Make your contribution as informative as is required (for the current purposes of the exchange).
	\item Do not make your contribution more informative than is required.
\end{enumerate}
The category \textbf{Quality} includes the supermaxim `\textit{Try to make your contribution one that is true}' with the two maxims:
\setlist{noitemsep}
\begin{enumerate}
	\setcounter{enumi}{6}
	\item Do not say what you believe to be false.
	\item Do not say that for which you lack adequate evidence.
\end{enumerate}
The category \textbf{Relation} is only defined by the maxim:
\begin{enumerate}
	\setcounter{enumi}{8}
	\item Be relevant.
\end{enumerate}
The formulation of what this entails is not straightforward for Grice, but he relates it to the focus on topics within a conversation, and how they can be reasonably shifted. 

Figure~\ref{fig:grice_adapted_to_convLM} on the left shows the Cooperative Principle summarized. Grice notes that this list is not necessarily complete, and he proposes ``Be polite'' as a possible addition. We follow up on this as a fifth category in the next Subsection~\ref{ch:3_grice_contribution}, where we try to operationalize these maxims as measurable features for neural generation models.

\subsection{Appropriateness as Conversational Cooperativity}
\label{ch:3_grice_contribution_appropriateness}

A neural conversation model should produce grammatical correct and not semantically malformed contributions. In the literature this is often referred to as \emph{fluent}. In the category \textbf{Manner} Grice proposed that a conversation has to be clear and understandable to allow for a meaningful exchange. We introduce \textbf{Fluency} as our first neural conversation model related category. Fluency of a model can be approximated conveniently by calculating perplexity on a test set. Perplexity measures the probability with which a model would have generated the test data (normalized for length). Under the assumption that the training data are formally (or grammatically) fluent, low perplexity should be a good indicator for the model's ability to account for the syntactic structure of the dialogues and generated utterances \shortcite{Serban2016BuildingED}. Therefore, we can assume that fluency is achieved, if the unconditional probability assigned to the contribution is high. With that definition, fluency is a property that can be approximated without taking the context $C_t$ into account:
\begin{equation}
	m_\textrm{fluency}(c_t|C_t) \sim f(c_t),
\end{equation}
with $f$ being the probability function realized by the model proper. We discuss fluency problems and survey interventions in Subsection~\ref{ch:3_grice_survey_fluency}.

According to Grice's category \textbf{Quantity} a contribution should neither be more nor less informative than necessary. We operationalize that as \textbf{Informativity} for neural conversation models and separate that into two dimensions: Genericity, and Repetitiveness. We define a contribution $c_t$ as generic, if it fits equally well to many contexts $C_t'$, which should be avoided:
\begin{equation}
	m_\textrm{inf,gen}(c_t|C_t) \sim \frac{1}{|\mathcal{C}|}\sum_{C_t'\in \mathcal{C}}\frac{f(c_t,C_t)}{f(c_t,C_t')},
\end{equation}
where $f$ is the model proper, and $\mathcal{C}$ is a set of randomly sampled other contexts. A constructed example would be the following contribution from $P_B$:

\ex. \label{ex:example_uninformative_a}
$P_a$: Do you like movies?\\
$P_b$:\,\#\,\textit{I don't know!}

It is a fluent reply, but not informative, as it is, presumably, an equally good reply in many contexts. There are not well established informativity metrics for neural conversation models in the literature. \shortciteA{li-etal-2020-dont} and \shortciteA{Welleck2020Neural} used token distributions and rare word counts as metrics in their experiments. Under the assumption that generic contributions contain generic (high frequent) tokens, this might be a way of quantifying informativity for conversation models. Informativity could also be measured by comparing the conditional and the unconditional contribution probability, as specific context should not make generic contributions more probable. \\
Repetitiveness, the second dimension of informativity, measures whether $c_t$ repeats information from $C_t$:

\begin{equation}
	m_\textrm{inf,2}(c_t|C_t) \sim f(c_t,C_t)
\end{equation}
where $f$ can be simple (sub)string match. If $c_t$ repeats content from $C_t$ no new information is contributed to the conversation. For example:

\ex. \label{ex:example_uninformative_b}
$P_a$: Pulp Fiction was released in 1994.\\
$P_b$:\,\#\,\textit{Did you know that the movie was released in 1994?}

Counting repeating tokens or $n$-grams is a way to detect repetition. However, one can repeat content without using the same tokens. Therefore, word embedding comparison metrics might be more robust. We discuss informativity problems and survey interventions in Subsection~\ref{ch:3_grice_survey_informativity}.

The Gricean category of \textbf{Quality} suggests contributing with true statements. Therefore, models need to have some representation of what is true. We operationalize this as \textbf{Consistency} towards additional knowledge. Recent approaches can be categorized into two main directions: adding additional facts by providing a knowledge base and giving the model an implicit personality by adding descriptive persona information. We define $c_t$ as consistent with a knowledge base if it represents one or more facts $f_i$ correctly:

\begin{equation}
	m_{\textrm{consistency,facts}}(c_t|C_t) \sim f(c_{t},f_i),
\end{equation}

and consistent with a given personality if $c_t$ is consistent with every $p_i$ of the personality description:

\begin{equation}
	m_{\textrm{consistency,persona}}(c_t|C_t) \sim f(c_{t},p_i).
\end{equation}

The function $f$ can be a neural model that predicts fact and opinion entailments, similar to natural language inference approaches. There are no precise and established metrics in the literature for consistency. Evaluation has to be done with humans, which makes it difficult to compare the different approaches. Additionally, human evaluations are expensive and time-consuming, limiting the number of experiments that researchers can do, such as hyperparameter searches. Natural language inference models fine-tuned on persona or fact and utterance pairs might be a solution, but need to be explored.

In the category \textbf{Relation}, Grice proposes that a dialogue contribution should ``be relevant'' and therefore follow some vaguely defined relevance criteria related to the subject of a conversation. We operationalize that as \textbf{Coherence}. We define $c_t$ as coherent if it has a pragmatic relation to $C_t$, specifically, to the dialogue history $D_t$:

\begin{equation}
	m_{\textrm{coherence}}(c_t|C) \sim f(c_t|c_{t-1}, ... c_1, o_{t}, ... o_1).
\end{equation}

A fluent, informative and consistent contribution that is not coherent given the dialogue context could be:

\ex. \label{ex:incoherent_dialogue}
$f_1$: Uma Thurman is an actress.\\
$P_a$:\,When was Pulp Fiction released?\\
$P_b$:\,\#\,\textit{Uma Thurman is a very popular actress.}

No automated metrics have yet been proposed that cover these complex notions of dialogue \shortcite{deriu_survey_2021}. Again, evaluation dialogue coherence requires manual evaluation, which causes the same problems mentioned previously. We discuss coherence problems and survey interventions in Subsection~\ref{ch:3_grice_survey_coherence}.

Not derived from the four proposed categories from Grice, but perhaps on the statement: ``\textbf{Be~polite!}'', a neural conversation model has to follow several rules (the \textit{social norm}) to not be offensive to the user of that system or any other third person in general. We propose that a contribution $c_t$ follows the social norm, if $c_t$ can be seen as non-offensive by the dialogue partner or by any other third person who knows $C_t$ and listens to $c_t$. Therefore, in addition to the full context $C_t$ a set of social rules $R$ is needed to determine good social norm: 

\begin{equation}
	m_\textrm{social norm}(c_t|C_t) \sim f(c_t, C_t, K, R).
\end{equation}

\shortciteA{dinan2021anticipating} categorized harmful contributions of neural conversation models into three different types: Generating $c_t$ that are harmful by themselves because they contain offensive language:

\ex. \label{ex:social_norm_violation_1}
$P_b$:\,\#\,\textit{I hate you!}\\

Generating $c_t$ as agreements to previous contributions $o_i$ that were harmful:
\ex. \label{ex:social_norm_violation_2}
$P_a$:\,I hate him!\\
$P_b$:\,\#\,\textit{I agree with you!}\\

And lastly, generating $c_t$ that contain inappropriate advices in safety-critical situations: 

\ex. \label{ex:social_norm_violation_3}
$P_a$:\,Should I jump down this cliff?\\
$P_b$:\,\#\,\textit{I think this is a good idea!}\\

We discuss social norm problems and survey interventions in Subsection~\ref{ch:3_grice_survey_social_norm}. We sum up the operationalization of Grice's categories in Figure~\ref{fig:grice_adapted_to_convLM} on the right-hand side.

\section{Research History and Problems}
\label{ch:3_grice_history}

Before we discuss the possible intervention points for neural conversation models in Section~\ref{ch:3_grice_intervention_points} and survey the literature that addressed the above demands in Section~\ref{ch:3_grice_survey} we briefly go over the history and encountered problems in that area first.

\shortciteA{ritter2011data} were the first who transferred the task of phrase-based machine translation to the field of dialogue generation. Instead of translating, they reformulated the task to transform an utterance into an appropriate reply. They identified two important challenges: First, unlike in bilingual machine translation, a reply is not semantically equivalent to the utterance. This leads to a wider range of possible responses, and being fluent is more challenging. Secondly, a reply has more unaligned words compared to bilingual pairs and often parts of the stimulus utterance are not referenced in the reply at all. To be fluent the model has to learn to reference correctly and to use different tokens in a clear and understandable way. They used the messaging platform Twitter~\footnote{\url{https://twitter.com}} to get utterance-reply pairs by scraping. They collected a total of 1.3 million pairs, such as:

\ex. \label{ex:ritter_et_al_twitter_data_example}
$P_a$ (utterance): I'm slowly making this soup ..... and it smells gorgeous! \\
$P_b$ (reply): I'll bet it looks delicious too! Haha \\
\small{(Data was taken from \shortciteA{ritter2011data})}

The authors observed that in a natural conversation utterance-reply pairs have strong structural relationships, as shown in the example above. Their task can be expressed in our formulation as:

\begin{equation}
	c_t = \textrm{most\_appropriate}_c(R(c|o_t)).
\end{equation}

At this point, the goal was to find an appropriate reply (given an utterance) and not to carry on a conversation. In their evaluation, replies generated by the statistical machine translation approach were preferred by humans over those yielded by information retrieval approaches; in 15\% of the cases, participants even preferred them over gold answers from the dataset.

From here many improvements were made regarding data sources and model capacities, and research was motivated by the success of being able to generate contributions that seem to be fluent in themselves (or at least grammatically appropriate). \shortciteA{sordoni-etal-2015-neural} argued that having one utterance (in our formalization, just $o_t$) is not sufficient for generating an appropriate response ($c_t$) in a longer exchange, as relevant information can be placed earlier in the context. Therefore, they created utterance triples by scraping Twitter again, for example:
\\
\ex. \label{ex:example_twitter_corpus_triples}
$P_a$ (context): because of your game ? \\
$P_b$ (message): yeah I'm on my way now \\
$P_a$ (reply): ok good luck ! \\
\small{(Data was taken from \shortcite{sordoni-etal-2015-neural})}

Their approach can be expressed as:

\begin{equation}
	c_t = \textrm{most\_appropriate}_c(R(c|c_{t-1}, o_t)).
\end{equation}

Being closer to a real conversation, but still having the property of a static conversation history. The new utterance $c_{t-1}$ in the dialogue history led them to develop a new model architecture, since the machine translation sequence-to-sequence models don't allow a differentiation between speakers. Their new recurrent language model consists of two separate \ac{rnn} cells to encode $c_{t-1}$ and $o_t$. They could show in a human evaluation that participants preferred replies generated with this approach over replies from the previous machine translation approach.

The first step towards variable sized dialogue histories was done by \shortciteA{Serban2016BuildingED} with the introduction of a new hierarchical \ac{rnn} structure. One \ac{rnn} cell is supposed to encode an utterance into a fixed-sized utterance representation, and another \ac{rnn} cell encodes these representations into the final hidden (fixed-sized) state. This way the architecture allows variable dialogue history lengths, which was an important step towards modern neural conversation models. However, in their experiments they still used triples for training by processing the Movie-DIC dataset \shortcite{banchs-2012-movie} into 245k three turn dialogue samples. 

\shortciteA{Serban2016BuildingED} also showed that pre-training greatly reduces model perplexity (a measure of quality of the model, where lower numbers are better; see below): They used 5.5 million utterance-reply pairs from the SubTle corpus \shortcite{Ameixa2014LukeIA} before fine-tuning and were able to improve perplexity from 35.63 to 27.09 on the static RNN model and to 26.81 on the hierarchical \ac{rnn} architecture. 

Later, \shortciteA{NIPS2017_3f5ee243} showed that architectures solely based on Self-Attention layers (proposed as the Transformer model) outperform recurrent neural networks in machine translation, and \shortciteA{radford2018improving} introduced the \ac{gpt} model (and their successor \ac{gpt}-2). Most recent approaches make use of these pre-trained Transformer-based models. In the ConvAI2 challenge \shortcite{DBLP:journals/corr/abs-1902-00098}, researchers were tasked to build the best conversation model on the PersonaChat dataset \shortcite{zhang2018personalizing}. The winning model by \shortciteA{wolf2019transfertransfo} was a \ac{gpt} model fine-tuned on the dialogue dataset. They trained and added additional embeddings to differentiate between the persona profiles and the dialogue history. The model reached a perplexity of 16.28 on the unseen test set, compared to 29.01 perplexity for sequence-to-sequence models based on recurrent neural networks. Transformer-based models have high capacity, but do not work well without the generative pre-training. However, with unsupervised pre-training they appear to learn the structure of language and word embeddings well, which makes them superior compared to the previous approaches.

The described improvements of language generation had an important impact on neural conversation models, since they appear to be fluent (or at least grammatically appropriate), which is important. But that alone is not sufficient for a good conversation, as discussed above. Lately, a large body of research has also accumulated investigating how to improve informativity, consistency, coherence and adherence to social norm, and not only by improving model architectures, but through interventions at a whole range of entry points around the base language model and along the processing pipeline. In the following section we go over these, as we call them, \textit{intervention points}.

\section{Intervention Points}
\label{ch:3_grice_intervention_points}

\begin{figure}
	\centering
	\includegraphics[scale=1]{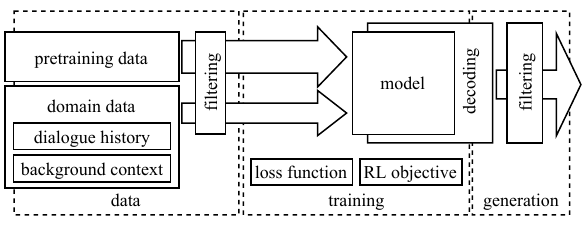}
	\caption{Intervention points for developing current non-goal driven neural conversation models.}
	\label{fig:grice_intervention_points}
\end{figure}

Figure~\ref{fig:grice_intervention_points} illustrates the schematic organization of the neural conversation models that are our topic here, in the context of both their training and their application. All the elements listed in the figure (and further described below) have been investigated as dimensions of change for improving the appropriateness of the productions of the model as a whole, as we will see in the next sections.

The core is formed by what we label here the \textit{model} proper, i.e., the conditional language model $P(c|C)$. These sequence-to-sequence architectures encode textual information (here dialogue history $D_t$ and optional private information $K$) into a hidden representation, which is transformed by a decoder into a new contribution $c_t$. They are either based on Self-Attention layers \shortcite{NIPS2017_3f5ee243} in various architectures or on \ac{rnn}s \shortcite{10.1162/neco.1997.9.8.1735,cho-etal-2014-properties}. They are designed to transform a sequence of fixed-sized input representations (here word embeddings) into a fixed-sized latent representation (encoder) or to generate fixed-sized output representations (here probability distributions over the vocabulary) based on a latent representation (decoder).

Although their task is to produce dialogue contributions, these models are typically pre-trained on (monological) text material, here labelled \textit{pre-training data}. Transformer-based architectures require pre-training but \ac{rnn}-based approaches also benefit from it, as shown by \shortciteA{Serban2016BuildingED}. Pre-training is supposed to train word embeddings and general language modeling abilities. This has been proven to be very effective for many NLP tasks \shortcite{DBLP:journals/corr/abs-1810-04805}. Thereafter, models were fine-tuned on dialogue data $D_t$ and (optionally) context $K$, often called \textit{domain data} in the literature. Given the way the required data is collected, namely automatic or crowdsourced and often without good curation practices, rule-based \textit{filtering} can be applied to remove material from which a model should not induce patterns.

The common machine learning approach is to optimize model parameter $\Theta$ on a tokenized corpus $\mathbf{X}=\{\mathbf{x}_1, \mathbf{x}_2, ..., \mathbf{x}_n\}$ by minimizing the \emph{loss function}, usually negative log likelihood. Using reinforcement learning is a rather new approach in neural dialogue modelling and makes possible full sequence (contributions) sampling and the use of non-differentiable loss functions, the \emph{reinforcement learning objectives}. 

\textit{Decoding} itself is an intervention point, as it has a high impact on the generated contribution. This is particularly the case for models trained with the language modelling loss, as this training regime cannot account for error accumulation when generating sequences. Decoding is not only about finding the best beams but also to control the generated sequence with different rules, as we will show in the next section.

Finally, the last intervention point we are discussing here is to apply a \textit{post-filter}. Based on specific rules, fully generated candidate contributions can be rejected or ranked to improve the conversation model.

\section{Survey of Proposed Interventions}
\label{ch:3_grice_survey}

Now, in this section, we survey approaches from the literature that tried to improve neural generation models regarding our proposed operationalized categories, sorted by the intervention points. For each category, we start with examples of problematic model output from the literature and then discuss how these problems have been addressed.

\subsection{Manner (Clarity) // Fluency}
\label{ch:3_grice_survey_fluency}

Language models can produce quite fluent utterances and \emph{fluency} improved over the years significantly, as described in our historical overview in Section~\ref{ch:3_grice_history}, through improvements in model size and architecture as well as the availability of larger training datasets. One additional issue that causes low fluency is that decoding strategies that optimize for sequences with high probability tokens (such as the original beam search algorithm) produce generic and repetitive output \shortcite{holtzman2020curious}. Language models tend to compute the highest probabilities to these sequences. They found that the probability for repeating phrases increases with each repetition, which creates a positive feedback loop and that text generated by humans has much lower probabilities, compared to text generated by beam search. This variance appears to be important in order to produce fluent output, and therefore different decoding strategies and various rules of thump were explored and are used in modern neural conversation models to reduce repetition and try to control sequence lengths.

\paragraph{Pre-training Data}
Model size and pre-training data are important factors to decrease perplexity and therefore increase the fluency. We already listed Important pre-training datasets in our historical overview in Section~\ref{ch:3_grice_history}.

\paragraph{Decoding}
Different decoding strategies prevent repetition by blocking repeating tokens or by adding randomness to that process. Randomness reduces the overall likelihood of the generated sequences and therefore repetition and since this is closely related to \emph{informativity}, we will discuss that in the next section. To maintain fluency it is important to block token-based repetition, as loops make utterances unfluent. \shortciteA{roller-etal-2021-recipes} evaluated decoding strategies in the context of neural conversation generation. To avoid loops they explicitly block tokens or n-grams that already occurred in the generated sequence. That means at every decoding step, regardless of the estimated token probabilities by the model, tokens that would cause a repeated 3-gram either in $o_t$ or $c_t$ are ignored.  The idea of beam blocking originally came from \shortciteA{paulus2018a} and can be applied to any decoding algorithm.

\paragraph{Objective Function}
The root cause of repetition might be the training objective itself, since assigning high probabilities for frequently occurring tokens is done by the model itself. Unlikelihood training \shortcite{li-etal-2020-dont} is a way of improving the objective function in several ways. It reduces the gap between training (next-token prediction only) and inference (generating full sequences) through an additional inverse loss $\lambda_{\textsc{UL}}$, that is added to the usual maximum likelihood estimation loss $\lambda_{\textsc{MLE}}$:

\begin{equation}
	\lambda_{\textsc{ULE}} = \lambda_{\textsc{MLE}} + \lambda_{\textsc{UL}},
\end{equation}
and is called the unlikelihood loss for step $t$ with a set of negative candidate tokens $C^{t}$:

\begin{equation}
	\lambda_{\textsc{UL}}(p_{\theta}(\cdot|x_{<t}), C^{t}) = - \sum_{c \epsilon C^{t}}{\textrm{log}(1 - p_{\theta}}(c|x_{<t})).
\end{equation}

The negative candidates $C^{t}$ can be set to anything that causes non-appropriate contributions. We will follow up on this approach in later sections as well. For \emph{fluency} it is important to avoid repetition. The authors therefore added negative candidates with repeating n-grams according to the following scheme:

\begin{equation}
	C_t^{\mathrm{context copy}} = 
	\begin{cases}
		\{y_t\} & y_t \epsilon \mbox{repeating context}\\ \emptyset &\mbox{otherwise,}
	\end{cases}
\end{equation}

where $y_t$ is a token in a repeating context n-gram. This should reduce the probability of tokens that form an n-gram that is already been decoded at inference.

\paragraph{Reinforcement Learning}
Reinforcement learning allows the use of non-differentiable functions and some are supposed to tackle repetition. \shortciteA{mesgar-etal-2021-improving} introduced a low repetition reward by counting occurrences of 1-grams to compute the following token ratio:

\begin{equation}
	\textrm{R}_{\textrm{rep}} = 1 - \frac{\textrm{\# repeated-tokens-in-response}}{\textrm{\# tokens-in-response}}
\end{equation}

The reward function measures repeated tokens and since the reward improves over the training, their approach seems to work. Unfortunately, they don't compare repetition between both models explicitly. The human evaluation focused on semantic plausibility and factual consistency, which we will discuss in future sections, and also didn't cover repetition. 

\subsection{Quantity // Informativity}
\label{ch:3_grice_survey_informativity}

Neural conversation models trained on large dialogue corpora with standard language modelling loss tend to generate ``simple, short and sometimes unsatisfying answers'' and predicting the next word might be an oversimplification \shortcite{vinyals-etal-2015-a-neural-conv}. Similarly, \shortciteA{li-etal-2016-diversity} observed that responses generated by generative conversation models with the highest average probability log-likelihoods tend to be trivial or generic, like \textit{I don't know!} or \textit{I'm OK}. The authors stated that models assign high probability to generic responses, if they are optimized for the likelihood, as they are suitable for many contexts. \shortciteA{Serban2016BuildingED} observed the same behavior and offered several explanations: Data scarcity could cause the models to only predict the most frequent utterances. The high  number of punctuation mark tokens in the data set makes it hard for the model to learn and predict diverse utterances, since the error signal is dominated by these frequent tokens. And the dialogue context could be too short, and more dialogue history is needed to diversify generated content.

More recently, \shortciteA{holtzman2020curious} observed that per-token probability of text created by humans is on average lower than text generated by language models, meaning that human-generated text is more surprising and models trained on minimizing negative log-likelihood on next-token prediction don't or cannot learn the real token distribution of human text. This problem is more general and not only tied to neural conversation models. The objective function, which only takes one true next-token into account, cannot capture the diversity of natural language or, specifically for this research area, the diversity of natural conversations. Prioritizing high probability tokens makes text less surprising and therefore less informative, as such probability distributions have lower entropy on average. Minimizing the cross-entropy loss causes the model to avoid making mistakes and because bland utterances fit to many contexts, it is less likely that a bland utterance is inappropriate. Current research focus on improving the objective function, better models and a lot on decoding algorithms.

\paragraph{Objective Function}
To improve informativity better objective functions are needed. Unlikelihood training \shortcite{li-etal-2020-dont,Welleck2020Neural} also contributes to informativity: the authors introduced a loss based on the frequency of occurring tokens: 

\begin{equation}
	\beta (y_c) = p_{\mathrm{model}}(y_c) \mbox{log}(\frac{p\mathrm{model}(y_c)}{p_{*}(y_c)}),
\end{equation}

with $p_{\mathrm{model}}(y_c)$ being the model's unigram distribution of the candidate $y_c$ and ${p_{*}(y_c})$ the pre-estimated human distribution. For these candidates, they use earlier contributions to explicitly punish repetition and tokens with large distribution difference to punish frequent (and therefore generic) tokens. This causes the negative candidate token probabilities to decrease and therefore reduces the probability of sequences with these candidates. 
\shortciteA{li-etal-2020-dont} reported a higher rare words cumulative mass with increasing unlikelihood objective scaling in their experiments using that approach compared to the usual language modelling approach. They further showed that this metric also correlates with human judgement (participants had to choose between two models through the question: ''\textit{Which response answers the question better?''}). Because of this, we see an indicator for token distribution characteristics being an important part of quantifying informativity. \shortciteA{Welleck2020Neural} observed similar success: They showed that using this loss in order to decrease probabilities of inappropriate tokens, it increases the number of unique tokens generated by the model on the test set from 11.8k to 13.8k. Compared to a human reference of 19.8k unique tokens, this is still significantly worse, but a promising step towards more informative contributions.

\paragraph{Model}
The \textsc{SpaceFusion} model by \shortciteA{gao-etal-2019-jointly} tries to increase relevance and interestingness. The latter correlates with informativity, since bland and generic content should be less interesting. Their idea is to spread the response distribution across a latent space to cluster them along diversity directions. In each direction, the distance from the true response additionally identifies the relevance. Two model parts are trained simultaneously: a sequence-to-sequence encoder (generating response representations $z_{\mathrm{s2s}}$) and an auto-encoder (generating target response representations $z_{\mathrm{AE}}$). Both share the same decoder. The loss is based on the interpolation $z_{\mathrm{interp}}=u z_{\mathrm{s2s}} + (1 - u) z_{\mathrm{AE}}$ with $u \sim U(0,1)$:

\begin{equation}
	\lambda_{\mathrm{interp}}(y) = - \frac{1}{|y|} \mbox{log}p(y|z_{\mathrm{interp}}).
\end{equation}     

That leads to semantically different responses being pushed into different directions in the latent space by using the shared decoder. The authors claim that their Space Fusion model was ranked highest in terms of relevance and interest compared to several baselines by a human evaluation.

\paragraph{Decoding}
Most research in terms of improving informativity is based on decoding, even though the decoding itself is not the core problem of it, as discussed earlier. Better decoding algorithms, however, that try to \emph{cover} the original problems by having specific rules or by adding randomness show improvements in this dimension.

Repetition does not only hurt fluency, but also informativity. \shortciteA{see-etal-2019-makes} experimented with a weighted decoding to tackle word and phrase repetition to reduce uninformative repeating utterances. They first identify repeating bigrams in the currently generated utterance and the previous utterances from the history, as well as repeating content words. Then, they apply a negative weight to these tokens. The authors used that repetition control in all experiments and highlighted the importance of it. However, this highlights the problematic model output (that seem to be repetitive) and does not address the real problem. If the weight is $-\infty$, that method is similar to the n-gram blocking explained in the previous section. This n-gram blocking also helps with informativity, as blocking n-grams from previous utterances helps to increase informativity through restricting the algorithm to repeat what was already said.

Another way of reducing repetition is to add randomness to the decoding algorithms. \shortciteA{holtzman2020curious} showed that  sampling one of the top k tokens (top-k-sampling) or setting the sampling threshold to a cumulative probability (top-p-sampling) is beneficial. Because of that, they come up with Nucleus sampling, a combination of both: From a token distribution $P(x|x_{1:i-1})$ the top-$p$ vocabulary $V^{(p)}$ (the set with the least number of tokens with a cumulative probability~$>= p$) is used to reshape that distribution to:

\begin{equation}
	P'(x|x_{1:i-1}) = 
	\begin{cases}
		P(x|x_{1:i-1})/P' &\mbox{if } x \epsilon V^{(p)}\\ 0 &\mbox{otherwise.}
	\end{cases}
\end{equation}

They stated that this approach gives more flexibility, as the sampling process is dynamic. However, \shortciteA{holtzman2020curious} and \shortciteA{galetzka-etal-2021-space} showed in their experiments that increasing randomness reduces consistency and coherence, since low probability tokens might cause contradictions when sampled.

Very short utterances are generally uninformative. Forcing decoding algorithms to create a minimum number of tokens by restricting the end-of-sequence sampling until a minimum sequence length is reached helps to create longer utterances, but is static. Sometimes short utterances are more suitable. \shortciteA{roller-etal-2021-recipes} expanded on this idea and proposed a model that predicts the sequence lengths of next utterances given the dialogue history. At inference, they use this estimation to set the minimum length for the decoding algorithm. While this helps to generate longer sequences when it is required to increase informativity, it does not solve the core issue, like the previous approaches for decoding.

\subsection{Quality // Consistency}
\label{ch:3_grice_survey_consistency}

Offering factual information is part of a natural conversation and (according to Grice) they should be correct. Neural conversation models solely trained on dialogue data have to have this explicit information encoded in its weights. Gathering such information from pure dialogues is a huge generalization challenge, and \shortciteA{Dinan2019WizardOW} observed that this information is indeed often wrongly remembered and used by the models. A common way is to induce knowledge into a model in addition to the dialogue history. This causes two problems: The train part now also includes the knowledge processing, which makes the training objective harder and the model architecture needs changes or gathering augmentations, as usually these models have a limited amount of input space for the additional knowledge. 

Even if supported with context $K$, neural conversation models tend to generate contributions that do not entail with $K$, which is called \emph{hallucination}. A recent survey \shortcite{ji2022survey} covered hallucination in the context of natural language generation in general. \shortciteA{xiao-wang-2021-hallucination} also investigated these phenomena and showed that there is a positive correlation with hallucination and predictive uncertainty. The predictive uncertainty for a token $x_i$ is measured by their entropy:
\begin{equation}
	H(x_i|C_i)=- \sum_{v \epsilon V} P(x_i=v|C_i) \mbox{log } p(x_i|C_i)
\end{equation}
with $C_i$ being the current context for $x_i$ and $V$ the set of all possible tokens (vocabulary).

Similar to the problem with factual consistency a neural conversation model only trained on dialogue also has an inconsistent personality: The datasets often contain many different personalities and as long as they are not explicitly labelled, this data is inconsistent by itself for the model \shortcite{li-etal-2016-persona}. Additionally, due to simple training objectives that do not cover personality explicitly, the model cannot generalize good enough to learn different and consistent behaviors \shortcite{vinyals-etal-2015-a-neural-conv}. Again, this is tackled by augmenting datasets with additional personality information along with new model architectures.

\paragraph{Domain Data}

Being consistent towards $K$ requires augmented dialogues for training. To our best knowledge, there are no large pre-training datasets of this type available, as they require numerous resources to make. Therefore, many fine-tuning datasets were built to learn consistent behavior. 

\shortciteA{Dinan2019WizardOW} introduced the knowledge augmented conversational dataset ``Wizard of Wikipedia''. In the collection process, one crowd-worker had access to Wikipedia excerpts that the was tasked to use while the other crowd-worker was tasked to ``\textit{play the role of a curious learner, eager to chat}''. They collected 22 thousand dialogues (over 200 thousand speaker turns) on a knowledge database with 93 million sentences across 1.4 thousand different topics. 

\shortciteA{zhang2018personalizing} proposed the personality-focused dataset PersonaChat. It consists of dialogues based on sets of 4--5 sentences (also crowdsourced) that describe a personality. Participants were tasked to get to know each other while pretending to be their given personality. The authors further asked another set of crowd-workers to rewrite the persona descriptions afterwards, since they noted a significant amount of word overlap between utterances and persona descriptions. This should emphasise the model to understand the personas rather than copying parts of it into the conversation.

Two additional important datasets are created by \shortciteA{li-etal-2016-persona}. The authors proposed dialogue datasets to address inconsistency towards model personality. The Twitter Persona dataset was extracted from Twitter in 2012 and only interactions where both participants had engaged in at least 60 3-turn conversations during the collection period were collected. This resulted in a dataset with labelled personalities (with the assumption that each participant has one consistent personality) with many data points per personality. They collected 74 thousand different writer sources with at least 60 different contributions to 3-turn conversations. Validation and test data were collected later in 2012 to increase the likelihood of uncorrelated data. The second dataset, Television Series Transcripts, had a focus on longer and more dialogues from the same personality but with a greatly reduced number of speakers. For that, they processed scripts from the television series Friends and The Big Bang Theory, which resulted in 69 thousand speaker turns across 13 different personalities (characters).

\shortciteA{galetzka-etal-2020-corpus} introduced the KOMODIS (Knowledgeable and Opinionated Movie Discussions) dataset with a focus on combining knowledge and personality, where crowd-workers were tasked to talk about different movies. They were provided with shared and private knowledge, as well as opinions (for personality) about the movies, persons and facts. The final dataset contains 7.5 thousand dialogues (100 thousand utterances) about 500 different movies.

\shortciteA{smith-etal-2020-put} also had the idea to combine knowledge and personality and created the Blended Skill Talk dataset. They used parts from the previously discussed datasets Wizard of Wikipedia, PersonaChat and Empathetic Dialogues \shortcite{rashkin-etal-2019-towards} to crowdsource new dialogues based on different contexts. Dialogues are seeded with a pair of utterances from one of the datasets (and some context) and two profile sentences from PersonaChat. Crowd worker can choose to freely talk or use a generated next utterance to prevent them from getting stuck in a specific skill. The generated utterances come from models fine-tuned on these datasets beforehand. They collected 5 thousand conversations with an average of 11.2 utterances.

A different dataset is the BeliefsBank dataset \shortcite{kassner-etal-2021-beliefbank} that consists of 2.6 thousand statements either labelled as false or true in the form of triples. Each triple consists of a sentence, a label and a strength value, for example: (``\textit{A poodle is a dog.}'', True, 0.9) or (''\textit{An eagle is a mammal}``, False, 1.0). This does not augment dialogues explicitly, but can be used independently to control or learn beliefs about the world.

\paragraph{Models}
Another large intervention point for consistency is the model itself. \shortciteA{wolf2019transfertransfo} proposed a simple but effective way of using a \ac{gpt} model without architecture changes: They augment the dialogue history with the context from PersonaChat and separated the persona profiles from the dialogue history by a segments layer. That layer is added to the normal word tokens and differentiates between the source using two new segment tokens. This way, the model learns to differentiate between persona description token embeddings and token embeddings from the dialogue history. The amount of additional context information in this approach is limited by the input sequence length of the Transformer models.

The following contributions focus on adding additional layers or memory modules to pre-trained language models. 

\shortciteA{zhao-etal-2020-knowledge-grounded} proposed a way to select a subset of relevant knowledge based on the dialogue history from a large knowledge base by a knowledge selection module utilizing a \ac{bert} model. Experiments are done with the Wizard of Wikipedia dataset. And since there are no knowledge subset labels available, they constructed weak labels automatically by assuming that gold answers have higher similarity with the correct context. These weak labels are only used to bootstrap the knowledge selection model, since random knowledge estimation at the beginning leads to error accumulation (feeding entirely wrong knowledge to the \ac{gpt}). After bootstrapping, the approach is unsupervised. A human evaluation indicates that the model is superior regarding context consistency and relevance compared to the proposed memory model by the authors of the dataset.

\shortciteA{lin-etal-2020-generating} experimented with a knowledge copy mechanism. They use \ac{lstm} cells for encoding the dialogue history, a Transformer to encode the knowledge and then another \ac{lstm} cell and a pointer network with a differentiable soft switch to either decode based on the history or based on the knowledge. Therefore, their model is able to either create text with standard language modelling (generating) or by explicitly copying tokens from the context. They also evaluate their approach on the Wizard of Wikipedia dataset. According to their human evaluation, this approach creates fluent and consistent dialogues, but they do not compare their approach to others.

\shortciteA{fan-etal-2021-augmenting} introduced a knowledge fetching based on \ac{knn} algorithms. Given a dialogue history $\mathbf{X}={\{x_1, x_2, ..., x_m\}}$ and context $\mathbf{E}={\{e_1, e_2, ..., e_m\}}$, the \ac{knn} algorithm identifies the closest information in $\mathbf{E}$ that is relevant to $x_i$. And since \ac{knn} is fully differentiable, they were able to incorporate this algorithm into a deep learning setup with a Transformer-based language model. A human evaluation showed that this architecture produces more consistent knowledge compared to retrieval-based approaches. Participants had to choose between both models, and around 60 \%  preferred the author's model.

\shortciteA{DBLP:journals/corr/abs-2107-07566} introduced a search engine augmented generation, an algorithm that estimates a search query for a black-box search engine and uses the results for generation. Their method is based on two components: A search query generator (an encoder-decoder Transformer) that uses the dialogue context to generate a search query and another encoder-decoder model that encodes the search results, concatenates them to the dialogue context and then generates the next response. Their human evaluation proofed the feasibility of this approach. Human annotators evaluated 750 responses across 100 model conversations and compared the search engine-based method with a no-knowledge baseline. The new approach outperformed the baseline in all metrics: more knowledgeable (46.5\% of the time; compared to 38.4\%), more consistent (76.1\% compared to 66.5\%) and more engaging (81.4\% compared to 69.9\%).

\paragraph{Objective Function}

Unlikelihood training (see Section~\ref{ch:3_grice_survey_fluency}) from \shortciteA{li-etal-2020-dont} also provides possibilities to explicitly train a model to be consistent: The unlikelihood loss $\lambda_{\textrm{UL}}$ penalises tokens that are part of inconsistent content. Here the challenge is to create inconsistent candidate pairs $(\mathbf{x}, \mathbf{y}^-)$. The authors refer to approaches from natural language inference tasks. They use an existing corpus of utterance pairs as well as utterance-persona pairs, semi-automatically annotated for entailment, contradiction or neutral relations. This approach is also suitable for tackling \textit{incoherence}, which we discuss in Section~\ref{ch:3_grice_survey_coherence}. 

Good knowledge selection mechanisms are important to create consistent content, as choosing the wrong knowledge in the first place eliminates the ability to create appropriate contributions at all. \shortciteA{chen-etal-2021-unsupervised} introduced a new distilled distant supervision loss to bootstrap correct labels for knowledge selection. The loss is a combination of a confidence score estimated by word overlap between knowledge candidates $k^i_t$ and gold response $y_t$:
\begin{equation}
	W_{k^i_t} = \mbox{softmax}(\frac{2 \cdot | \mbox{set}(k^i_t) \cap \mbox{set}(y_t)|}{|\mbox{set}(k^i_t)| + |\mbox{set}(y_t)| + \epsilon})
\end{equation}

and a teacher module (that has access to the gold response) that is used to transfer the knowledge to the student (the real) module, which aims to reduce the noise in the labels, which they call knowledge distillation $\lambda_{\mathrm{KD}}$. The final loss is:
\begin{equation}
	\lambda_{\mathrm{DDSL}} = \lambda_{\mathrm{CE}}(S,k^{'}_t) \cdot W_{k^i_t} + \lambda_{\mathrm{KD}}
\end{equation}
with $\lambda_{\mathrm{CE}}(S,k^{'}_t)$ being the cross entropy loss between a knowledge selection distribution $S$ and estimated knowledge $k^{'}_t$.
They reported improvements in knowledge selection for their unsupervised setup over previous supervised setups, including the baseline for the Wizard of Wikipedia dataset.

\paragraph{}

Another intervention point to tackle inconsistency is reinforcement learning \shortcite{mesgar-etal-2021-improving} (see Section~\ref{ch:3_grice_survey_fluency}): Similar to unlikelihood training, the authors use an \ac{nli} model that can determine entailment between a dialogue utterance and context. For that, they fine-tuned a BERT model with the Dialogue NLI dataset \shortcite{welleck-etal-2019-dialogue} (an adaption of the PersonChat dataset) to predict whether a generated utterance entails or contradicts a personality trait. The classification probabilities of the model are then used to compute a persona consistency reward:  
\begin{equation}
	R = \frac{1}{|p|} \sum_{f_i \epsilon p} P^{\mathrm{NLI}}_e(f_i, r) - \frac{\beta}{|p|} \sum_{f_i \epsilon p} P^{\mathrm{NLI}}_c (f_i, r)
\end{equation}
used to evaluate fully sampled sequences $r$. The reward sums up the entailment probabilities $P_e^{\mathrm{NLI}}$ and the contradiction probabilities $P_c^{\mathrm{NLI}}$ for all personality traits $f_i$ compared to $r$. $\beta \geq 1.0 $ as a hyperparameter is used to slightly penalize contradictions more over entailments.

\subsection{Relation // Coherence}
\label{ch:3_grice_survey_coherence}

The discrepancy between standard language modelling loss and the real communicative intent of a human conversation (mentioned in Section~\ref{ch:3_grice_survey_informativity}) also leads to incoherent responses \shortcite{vinyals-etal-2015-a-neural-conv}. \shortciteA{zhao-etal-2017-learning} stated further that simple sequence-to-sequence models cannot acquire the ability to generalize over situational context without explicit guidance. This is further hindered by having many good contributions to a dialogue state, but only learning a specific one (the gold next utterance). Another problem is that models do not make use of the full observable dialogue history $C_t$: \shortciteA{sankar_neural_2019} studied the sensitivity of recurrent and Transformer-based dialogue models to synthetic perturbations $D^P_t$ of $D_t$ (order of words and utterances, dropping certain words or utterances) using four different dialogue datasets. They showed that the perplexity difference between $D_t$ and $D^P_t$ is small, concluding that these models are insensitive to order, which is problematic for capturing conversational dynamics. Many approaches across different intervention points have been proposed recently.

\paragraph{Pre-training}
\shortciteA{mehri-etal-2019-pretraining} did not introduce new pre-training data, but several pre-training tasks to create strong dialogue representations for down stream tasks like next utterance generation or dialogue act prediction. To model coherence, they proposed the task of \ac{ini}: the objective is to determine a wrong (randomly chosen) utterance, which replaced a correct utterance in the dialogue history. They reported that this task improves on F1 and \ac{bleu} on their test set. 

\paragraph{Objective Function}
\shortciteA{zhou-etal-2021-learning} integrated the perturbations $D^P_t$ into the train process by defining a reward function based on whether $D^P_t$ has an effect on the likelihood of the next utterance or not:
\begin{equation*}
	R(Y |X, X') = NLL_{adv} - NLL_{orig}
\end{equation*}
Additionally, they defined a penalty function using a (not further specified) margin $\mathcal{M}$:
\begin{equation*}
	P(Y |X, X') = min(0, NLL_{adv} - NLL_{orig} - \mathcal{M}).
\end{equation*}

They pre-trained their model with the standard language modelling objective until perplexity on validation converges, and then continued the training with their inverse adversarial method, where the next utterance is alternately taken from the training set or generated by the model itself. A human evaluation in terms of fluency, consistency (not our definition, but asking: ``how likely the generated text is related to the input dialogue history''), and diversity showed that this approach improves greatly over a simple baseline and also outperformed other approaches in all three categories. 

\paragraph{Model}
The model itself is the most researched intervention point for maintaining coherence. \shortcite{zhao-etal-2017-learning} introduced approaches based on \ac{cvae} for dialogue generation to account for the complexity of several possible continuations for a given context. The model encodes the dialogue context (together with meta information on the topic) as a distribution vector to allow for variation in the decoding step, where the original input is reconstructed from the latent representation. The distribution of the latent representation is learned by minimizing the \ac{kl} Divergence to a recognition network that has access to the target response as well. The model is trained on the SwitchBoard corpus \shortcite{godfrey_john_j_switchboard-1_1997}. They reported best results in terms of automatic metrics (\ac{bleu} and GloVe-based embedding distance) and a manual analysis of model outputs. \shortciteA{xu_better_2018} built upon this further by adding the calculated coherence score between the context and the gold response as an additional latent variable instead of dialogue act labels. Their results on the OpenSubtitles corpus suggest that this approach yields improvements, even when trained on the unfiltered data and evaluated on the filtered test set. \shortciteA{gu2018dialogwae} also adapted the \ac{cvae} idea by changing the training objective to a generative adversarial network discriminator to allow for more complex distributions in the latent space. This architecture enables the model to capture the multimodal distribution of possible responses and shows improvements in terms of the automatic metrics proposed by \shortciteA{zhao-etal-2017-learning}. Further, \shortciteA{dai-etal-2021-apo} showed that moving from euclidean to hyperbolic space for the latent representation yields further improvements in the direct comparison with \shortciteA{gu2018dialogwae}.

Increasing informativeness can sacrifice dialogue coherence \shortcite{gao-etal-2019-jointly}. Both can be optimized jointly by using the already proposed SpaceFusion model. This approach improves over baselines and the \ac{cvae} model from \shortciteA{zhao-etal-2017-learning} in \ac{bleu} and human rated relevance and interest.

\shortciteA{bao-etal-2020-plato} proposed another model architecture based on conditional training, which they call PLATO. They jointly trained a latent speech act classifier and a response generator. Their model is trained unsupervised and also includes the bag of words loss introduced by \shortciteA{zhao-etal-2017-learning} to learn to reconstruct the response words from the latent speech act's hidden state $z$. During inference, one response is generated for all values of $z$ and a response selection component is used to find the highest scoring (i.e., most coherent) response. Their human evaluation on 100 random samples, where judges were asked to rate coherence as ``whether the generated response is relevant with the dialogue context and consistent with the expressed information or background knowledge" on a three-point Likert scale, showed improvements across all categories. Later, they proposed PLATO-2 \shortciteA{bao-etal-2021-plato} which is pre-trained on a cleaned Reddit corpus. It consists of a diverse response generation model based on their baseline approach and a separate response coherence estimator which combines the response selection loss from above with a masked language modelling objective. During inference, as above, the first model will generate a response for every latent speech act and the second one will select the highest scoring one. PLATO-2 models outperformed their counterparts (PLATO \shortcite{bao-etal-2020-plato}, DialoGPT \shortcite{zhang-etal-2020-dialogpt}, Blender \shortcite{roller-etal-2021-recipes} and Microsoft XiaoIce \shortcite{zhou-etal-2020-design}) in multiple human evaluations.

Referenced-based word overlap metrics such as \ac{bleu} do not correlate well with coherence \shortcite{lowe-etal-2017-towards}. To create appropriate training signals, one line of research focuses on models that are able to predict coherence scores. \shortciteA{lowe-etal-2017-towards} collected a dataset with human judgements on next utterance appropriateness to train a model with it. However, generic (and therefore uninformative) responses were often rated as highly appropriate, which again shows a negative correlation between informativity and coherence. \shortciteA{tao_ruber_2018} proposed a \ac{ruber}: The referenced metric $s_R$ calculates the similarity of a generated response and the ground truth with a cosine distance, and the unreferenced metric $s_U$ scores the appropriateness of a reply for a given context. This is done by training a neural network to distinguish between the gold next response $r$ and randomly sampled $r^-$ by minimizing:
\begin{equation*}
	\mathcal{L} = max(0,\Delta - s_U(q, r) + s_U(q, r^-))    
\end{equation*}
where $\Delta$ is a constant based on the development set. \shortciteA{ghazarian-etal-2019-better} extended this measure by using \ac{bert} embeddings instead of word-to-vec and some model modifications. A direct comparison with the \ac{ruber} metric showed that this measure can profit from the richer representations provided by \ac{bert}. \shortciteA{huang_grade_2020} noted that the above metrics only focus on the utterance level without taking topic transitions into account. To address this problem, they proposed to use \ac{grade}, which is based on sampled responses that are similar to the ground truth rather than sampled randomly.

\paragraph{Decoding}

\shortciteA{see-etal-2019-makes} suggested to use weighted decoding to control for different response attributes by changing the score of a possible next word according to specific weighted features $w_i$: 
\begin{equation*}
	\begin{split}
		score(w, y_{<t}; x) = score(y_{<t}; x) + log P_{CM}(w|y_{<t}, x) +  \sum_{i} w_i * f_i(w; y_{<t}, x)
	\end{split}
\end{equation*}
For response relatedness, they defined $f$ to be a cosine-based similarity measure of GloVe-based embeddings of the candidate next word and the last observed utterance $o_t$:\\
\begin{equation*}
	\begin{split}
		resp\_rel(w; y_{<t}, x) = cos\_sim(word\_emb(w), sent\_emb(o_t))
	\end{split}
\end{equation*}
This seemed to improve the topic relatedness of the response in a manual evaluation by the authors. However, this effect was not reproducible in their crowdsourced human evaluation, where workers were asked to judge the dialogue quality in terms of \textit{listening} and \textit{making sense} (among other qualities) after six turns of interactive chat. The authors attributed this discrepancy to the problem that deviations from the learned distribution like this can also lead to uncontrolled nonsensical output by the model. 
This confirms the findings of \shortciteA{baheti-etal-2018-generating}, who used a similar approach that combines response relatedness with an additional topic similarity feature for weighted decoding. They extracted the topic distributions for the context and the next utterance from a pre-trained HMM-LDA topic model and computed their similarity. In their human evaluation, they showed that an increase in content richness (similar to our definition of Informativity) comes with a decrease in plausibility (i.e., coherence). However, a combination of their approach with MMI re-ranking and beam search led to the best results in both categories (though still leaving quite some room for improvement).

\subsection{Social Norm}
\label{ch:3_grice_survey_social_norm}

The problem of generating contributions that do not follow the social norm is similar to the coherence problem from the previous section. While directly insulting contributions (for example: ``I hate you!'') are often caused by the observed language from the pre-training and domain data, the problem of agreeing to unsocial utterances and the problem of giving unsafe advice requires semantic understanding of the full context $C_t$ and the social norm $R$ \shortcite{dinan2021anticipating}. Since $R$ is not defined explicitly, the models also have to learn a social behavior or ethical understanding implicitly.

\paragraph{Pre-Training Data}
\shortciteA{dinan2021anticipating} further stated that harmful contributions like in Example~\ref{ex:social_norm_violation_1} can be avoided by cleaning up the train data, since datasets scraped from the web usually have a biased hegemonic world-view and tend to misrepresent social movements \shortciteA{10.1145/3442188.3445922}. Similarly, \shortcite{DBLP:journals/corr/abs-2010-07079} showed that models trained on data from Reddit with more unsafe contexts generate more unsafe utterances in a dialogue than models trained on the less toxic BlendedSkillTask \shortcite{smith-etal-2020-put} dataset. They measured toxicity as frequency of occurring toxic tokens in the generated text.

\paragraph{Domain Data}
\shortciteA{DBLP:journals/corr/abs-2106-10328} experiment with smaller datasets and fine-tuned a language model according to different behaviour types. They showed that fine-tuning with a set of handcrafted question answering pairs (``\textit{What makes a person beautiful?}'', ``\textit{The attractiveness of a person is a highly subjective measure.} [...]'') reduced toxicity measured in a human evaluation, which matches the statement from \shortciteA{dinan2021anticipating}. However, all these approaches do not help with indirect harmful content like in Example~\ref{ex:social_norm_violation_2} or unethical contributions like in Example~\ref{ex:social_norm_violation_3}.

\paragraph{Decoding}
Manipulating the decoding algorithms is a common intervention point to reduce toxicity, however, most of these approaches are in the context of general language generation. We still review two of them here, as these strategies could be adapted to a dialogue context as well. \shortciteA{schick2020self} introduced a self-diagnosis approach: They created sentence patterns like: ``Does the above text contain toxicity?'' and added them at the end of a text they want to analyse. Measuring the probability of `yes' and `no' tokens as a next token, seems to give a good toxicity estimation for bigger models. They report an accuracy of up to 87.3\% on the largest T5 model. With this self-diagnosis capacity they detect inappropriate tokens at decoding to avoid them as a next token. \shortciteA{liu-etal-2021-dexperts} experimented with non-toxic expert and (toxic) anti-expert language models. In order to create these models they fine-tuned a GPT-2 model on the dataset from the `Jigsaw Unintended Bias in Toxicity Classification' challenge.\footnote{\url{https://www.kaggle.com/c/jigsaw-unintended-bias-in-toxicity-classification}} The toxic model is only trained on toxic data, while the expert model is only trained on clean data. At decoding they look at tokens with a large probability difference (assuming that they are toxic) and modify the token probability distribution according to the measured toxicity while decoding.

\paragraph{Post Filtering}

Instead of encouraging a neural conversation model to stick to the social norm, this intervention point prevents the model from inappropriate contributions by denying already generated contributions. \shortciteA{dinan-etal-2019-build} introduced a four-step approach to train a robust detection system: Build an initial model on the Wikipedia Toxic Comments dataset that detects single-turn utterances. Then ask crowd workers to find contributions that are not covered by the model. Retrain the classifier and repeat the crowd-working task again. Since single-turn analysis is not sufficient, they also experimented with crowdsourced multi-turn samples and reached an F1 score of 66.4 on their best multi-turn classifier. \shortciteA{DBLP:journals/corr/abs-2010-07079} followed up on this work, but focused on eliciting offensive messages from the neural conversation model in the break-it phases. For that, they formulated four different strategies: training classifiers for detecting unsafe messages, making unsafe content explicitly unlikely, avoiding topics like politics or religion and forcing the model to respond gender-neutral. They reported an F1 of 85.9 on the same multi-turn data, and humans rated their best model as safe in 96,6\% of the time.

\section{Conclusion}
\label{sec:conclusion}

In this paper, we surveyed current problems and shortcomings with neural generation models for non-goal oriented dialogue. For that we took inspiration from Grice's Cooperative Principle and introduced five categories (fluency, informativity, consistency, coherence and social norm) that need to be followed to achieve a good conversation model. We analysed current solution proposals, sorted them according to our categories and possible intervention points, and highlighted their potentials and limitations.

Pre-trained language models are strong tools in the field of neural generation models, but unsupervised training on pure dialogues can only solve one aspect of a good artificial dialogue partner: they are able to make dialogues fluent. All other aspects require labelled high quality datasets and strong automated metrics for training and evaluation. We noticed an emerging trend away from pure end-to-end training (i.e., building chatbots) to more specific solutions for different dialogue aspects, such as diversification of generated words, learning the utilisation of general knowledge or using consistent personality traits. We support that trend, as human language and specifically dialogue, has a high complexity, and therefore the generalisation effort that neural networks have to make is also high. 

We want to stress one particular issue in the field, namely the lack of comparability: The aforementioned trend to very specific solutions leads to many different datasets that are used to train models and that are not comparable. Additionally, the lack of available metrics enforces the use of human evaluations or self-defined metrics, which are also not comparable among each other. That makes it hard to quantify the reining effect of different approaches, therefore comparability should receive more attention in future work.

\vskip 0.2in
\bibliography{anthology,acl2021}
\bibliographystyle{theapa}

\begin{acronym}
	\acro{rnn}[RNN]{Recurrent Neural Network}
	\acro{nli}[NLI]{Natural Language Inference}
	\acro{bert}[BERT]{Bidirectional Encoder Representations from Transformers}
	\acro{gpt}[GPT]{Generative Pre-trained Transformer}
	\acro{lstm}[LSTM]{Long Short-Term Memory}
	\acro{knn}[KNN]{K-Nearest Neighbours}
	\acro{ini}[InI]{Inconsistency Identification}
	\acro{bleu}[BLEU]{BiLingual Evaluation Understudy}
	\acro{cvae}[CVAE]{Conditional Variational Auto Encoder}
	\acro{kl}[KL]{Kullback Leibler}
	\acro{ruber}[RUBER]{Referenced metric and Unreferenced metric Blended Evaluation Routine}
	\acro{grade}[GRADE]{Graph-enhanced Representations for Automatic Dialogue Evaluation}
\end{acronym}

\end{document}